%% file: main.tex
\documentclass{article}

\PassOptionsToPackage{numbers, compress}{natbib}

    \usepackage[final]{neurips_2024}

\usepackage[utf8]{inputenc} 
\usepackage[T1]{fontenc}    
\usepackage{hyperref}       
\usepackage{url}            
\usepackage{booktabs}       
\usepackage{amsfonts}       
\usepackage{nicefrac}       
\usepackage{microtype}      
\usepackage{xcolor}         

\usepackage{graphicx}
\usepackage{booktabs}
\usepackage{bm}
\usepackage{amsmath}
\usepackage{footmisc}
\usepackage[capitalize]{cleveref}
\newcommand{\Eq}[1]{Eq.~(\ref{eq:#1})}
\newcommand{\Fig}[1]{Figure~\ref{fig:#1}}
\newcommand{\Table}[1]{Table~\ref{tbl:#1}}
\newcommand{\Sec}[1]{Section~\ref{sec:#1}}

\usepackage{CJKutf8}
\newcommand*{\Ja}[1]{%
  \begin{CJK}{UTF8}{ipxm}#1\end{CJK}
}
\usepackage{multirow}
\usepackage{array}

\title{
JRadiEvo: A Japanese Radiology Report Generation Model  Enhanced by Evolutionary Optimization of Model Merging
}

\author{%
    Kaito Baba,
    Ryota Yagi,
    Junichiro Takahashi,
    Risa Kishikawa,
    Satoshi Kodera\\[2mm]
Department of Cardiovascular Medicine\\
The University of Tokyo Hospital, Tokyo, Japan\\[2mm]
  \texttt{baba-kaito662@g.ecc.u-tokyo.ac.jp} \\
}

\begin{document}

\maketitle

\input{sec/abs}

\input{sec/01}

\input{sec/02}

\input{sec/03}

\input{sec/04}

\section{Conclusion}

In this study, we proposed a Japanese Radiology report generation model enhanced by Evolutionary optimization of model merging (JRadiEvo), marking the first attempt to extend a multimodal vision-language model for non-English medical text generation using evolutionary model merging. Despite utilizing only 50 translated samples from publicly available data, JRadiEvo demonstrated superior performance compared to leading models from recent studies trained on much larger datasets. This highlights the effectiveness of our approach in efficiently leveraging limited data to create a powerful and practical medical foundation model.

While JRadiEvo has shown promising results in evaluation metrics, human judgment by medical experts or further refinement may be needed to make it suitable for clinical use. Future work includes efforts to close this gap to ensure the model's reliability in real-world medical settings.

\section{Acknowledgments}
This work was supported by Cross-ministerial
Strategic Innovation Promotion Program (SIP) on ``Integrated Health Care System''
Grant Number JPJ012425.

\newpage
\bibliographystyle{IEEEbib}
\bibliography{refs}

\end{document}

%% file: sec/abs.tex
\begin{abstract}
With the rapid advancement of large language models (LLMs), foundational models (FMs) have seen significant advancements. Healthcare is one of the most crucial application areas for these FMs, given the significant time and effort required for physicians to analyze large volumes of patient data. Recent efforts have focused on adapting multimodal FMs to the medical domain through techniques like instruction-tuning, leading to the development of medical foundation models (MFMs). However, these approaches typically require large amounts of training data to effectively adapt models to the medical field. Moreover, most existing models are trained on English datasets, limiting their practicality in non-English-speaking regions where healthcare professionals and patients are not always fluent in English. The need for translation introduces additional costs and inefficiencies.
To address these challenges, we propose a \textbf{J}apanese \textbf{Radi}ology report generation model enhanced by \textbf{Evo}lutionary optimization of model merging (JRadiEvo). This is the first attempt to extend a non-medical vision-language foundation model to the medical domain through evolutionary optimization of model merging. We successfully created a model that generates accurate Japanese reports from X-ray images using only 50 translated samples from publicly available data. This model, developed with highly efficient use of limited data, outperformed leading models from recent research trained on much larger datasets. Additionally, with only 8 billion parameters, this relatively compact foundation model can be deployed locally within hospitals, making it a practical solution for environments where APIs and other external services cannot be used due to strict privacy and security requirements.
\end{abstract}


%% file: sec/01.tex
\section{Introduction}

In recent years, foundational models (FMs) have seen remarkable advancements, transforming various fields by offering more sophisticated and powerful solutions~\cite{bommasani2021opportunities}. A key driver of this progress has been the rise of large language models (LLMs), which have greatly expanded the capabilities of FMs, particularly in processing and generating text with high accuracy and contextual understanding. This has sparked exponential growth in research~\cite{zhao2023surveylargelanguagemodels}, leading to the development of vision-language models that integrate visual and textual data~\cite{gpt4v, NEURIPS2023_6dcf277e,NEURIPS2022_960a172b}, 
as well as fine-tuning approaches that enhance model performance for specific tasks~\cite{peng2023instruction,NEURIPS2023_5abcdf8e}.

Healthcare is one of the most critical application areas for foundational models. The need to develop models tailored to healthcare is essential, particularly because physicians often face the challenge of reviewing large volumes of medical data, such as X-rays, which can be time-consuming and demanding. Advanced FMs can help alleviate this burden by enabling quicker and more efficient diagnoses, improving the overall effectiveness of healthcare delivery and patient outcomes. In response to this need, various FMs have been fine-tuned specifically for the healthcare domain, further enhancing their accuracy and effectiveness in clinical settings~\cite{NEURIPS2023_5abcdf8e,singhal2023expertlevelmedicalquestionanswering,pmlr-v225-moor23a}.

However, despite these advancements, several challenges remain. One significant issue is that most of the models developed so far, such as LLaVA-Med~\cite{NEURIPS2023_5abcdf8e} and MedPaLM 2~\cite{singhal2023expertlevelmedicalquestionanswering}, are predominantly in English, whereas many healthcare professionals and patients are not always proficient in English. For these models to be truly practical, there is a pressing need to expand their capabilities to non-English languages. Relying on a two-step process, where the model first generates output in English and then translates it, can introduce additional costs and complexity, making it less efficient and accessible. Additionally, publicly available datasets that can be used to train these models, such as MIMIC-CXR~\cite{MIMIC-CXR} and IU X-Ray~\cite{iuxray}, are overwhelmingly in English, with very few datasets available in other languages. Translating the large amounts of data needed for training into other languages with high quality is a costly and resource-intensive process. This scarcity of non-English datasets makes it difficult to develop models that can handle non-English languages. Furthermore, due to privacy concerns, it is challenging to collect and use patient data for model training, further complicating the creation of such datasets. Also, the use of large models through APIs, such as GPT-4~\cite{openai2024gpt4technicalreport}, is often impractical in healthcare settings because of the stringent privacy regulations that protect patient data, which limits the deployment of these models in real-world clinical environments.

To address these challenges, this paper presents a \textbf{J}apanese \textbf{Radi}ology report generation model enhanced by \textbf{Evo}lutionary optimization of model merging~\cite{akiba2024evolutionaryoptimizationmodelmerging} (JRadiEvo), a first attempt to extend a multimodal vision-language model for non-English medical text generation by utilizing evolutionary optimization of model merging~\cite{akiba2024evolutionaryoptimizationmodelmerging}. JRadiEvo was developed by merging a non-medical vision-language model, medical text-to-text models, and a Japanese-language text-to-text model using an evolutionary algorithm. This innovative approach enabled the efficient creation of a Japanese radiology report generation model using only a minimal amount of Japanese-language data, addressing the critical need for non-English medical text generation in a resource-constrained environment.

Below we outline our key contributions, which aim to advance the field of multimodal foundational models in healthcare:

\begin{enumerate}
\item \textbf{Efficient use of limited non-English medical data}: In the context of the difficulty in collecting non-English datasets, JRadiEvo demonstrates the ability to create a non-English medical report generation model by translating and utilizing only 50 cases from publicly available English datasets. This approach highlights the efficiency of the development process, demonstrating how a non-English medical report generation model can be created using extremely limited data and annotations. Additionally, it is noteworthy that not only was the dataset used after translation limited to 50 cases, but the entire dataset used to create JRadiEvo consisted of just 50 cases. This underscores the fact that JRadiEvo efficiently utilizes a very limited amount of data, demonstrating an effective approach to handling medical data under strict privacy and security constraints.
\item \textbf{Novel application of model merging in the medical vision-language model}: Traditionally, adapting models to the medical domain has relied on fine-tuning or training from scrach. To the best of our knowledge, there are no existing study of applying model merging alone to adapt a vision-language model to the medical domain. While recent research~\cite{akiba2024evolutionaryoptimizationmodelmerging} has proposed using evolutionary optimization of model merging for vision-language models, this approach has been limited to natural images. To our knowledge, no prior studies have extended this technique to medical images or other domain-specific imagery beyond natural images.
\item \textbf{Lightweight model for local deployment}: JRadiEvo is an 8B parameter model, making it lightweight enough to be deployed on local hospital computing systems without the need for external APIs. This local deployment capability addresses critical privacy and security concerns, allowing hospitals to maintain control over patient data. Additionally, given the challenges of equipping facilities with expensive GPUs proportional to patient numbers, JRadiEvo's compact size and low GPU memory requirements make it practical for widespread use.
\item \textbf{Cost-efficient training process}: JRadiEvo eliminates the need for computationally expensive backpropagation during training, enabling a far more efficient learning process compared to training a new model or fine-tuning. Additionally, by leveraging model merging instead of fine-tuning, JRadiEvo avoids the common issue of catastrophic forgetting~\cite{MCCLOSKEY1989109,kotha2024understanding,luo2024empiricalstudycatastrophicforgetting} that often occurs during fine-tuning, allowing for a more stable and efficient development process.
\end{enumerate}

%% file: sec/02.tex
\section{Related work}

\subsection{Foundation models}

\paragraph{Medical large language models}
In the medical domain, several large language models (LLMs) have been developed and fine-tuned to achieve high performance. Notable models include ChatDoctor~\cite{ChatDoctor}, DoctorGLM~\cite{Xiong2023DoctorGLMFY}, BioGPT~\cite{10.1093/bib/bbac409}, Med-Alpaca~\cite{han2023medalpacaopensourcecollection}, PMC-LLaMA~\cite{wu2023pmcllamabuildingopensourcelanguage}, Med-Gemini~\cite{saab2024capabilitiesgeminimodelsmedicine}, Med-PaLM~\cite{medpalm}, and Med-PaLM 2~\cite{singhal2023expertlevelmedicalquestionanswering}. These models have demonstrated impressive capabilities in understanding and generating medical text by leveraging the power of LLMs fine-tuned for healthcare-specific tasks.

\paragraph{Vision-language models}
As for multimodal vision-language models (VLMs), prominent models like Flamingo~\cite{NEURIPS2022_960a172b}, Coca~\cite{yu2022coca}, BLIP~\cite{pmlr-v162-li22n}, PaLI-X~\cite{chen2023palixscalingmultilingualvision}, CogVLM~\cite{wang2023cogvlm}, GPT-4V~\cite{gpt4v}, and LLaVA~\cite{NEURIPS2023_6dcf277e} have been developed to integrate visual and textual data, pushing the boundaries of what can be achieved in multimodal learning.

\paragraph{Medical vision-language models}
Recently, there has been growing interest in extending these vision-language models to the medical domain, leading to the development of models like XrayGPT~\cite{thawakar-etal-2024-xraygpt}, MedFlamingo~\cite{pmlr-v225-moor23a}, Med-PaLM M\cite{tu2023generalistbiomedicalai}, LLaVA-Med~\cite{NEURIPS2023_5abcdf8e}, and CheXagent~\cite{chen2024chexagent}. These models incorporate LLMs into vision-language frameworks specifically designed for medical applications. LLaVA-Med~\cite{NEURIPS2023_5abcdf8e} is a vision-language model specifically designed for the medical field by using instruction-following data generated with GPT-4~\cite{openai2024gpt4technicalreport} to perform instruction-tuning on the LLaVA~\cite{NEURIPS2023_6dcf277e} model. CheXagent~\cite{chen2024chexagent} represents a significant advancement in medical vision-language models. It constructs a large-scale instruction-tuning dataset by aggregating publicly available datasets and adding new labels to existing ones, enabling accurate chest X-ray interpretation and showcasing the potential of instruction-tuned vision-language models in medical imaging.

Despite the advancements of these models, most rely on fine-tuning or training from scratch, which requires large datasets. In the medical field, where privacy and security concerns make it difficult to create large datasets, this dependency poses a significant challenge. While several publicly available English datasets can be used for training, there are very few non-English datasets, making it difficult to develop practical models for non-English-speaking regions. JRadiEvo addresses this issue by proposing an efficient method for creating a vision-language model in a non-English language using just 50 translated examples from a public dataset. Note that not only were 50 examples translated, but the entire medical image-text dataset for creating JRadiEvo consisted of only 50 cases. Furthermore, while existing models output text in English, JRadiEvo generates reports directly in Japanese, eliminating the need for translation and demonstrating the potential for practical use in non-English-speaking regions.

\subsection{Model merging}

Model merging is a technique that allows the strengths of multiple pre-trained models to be combined without the need for additional training. One prominent approach involves using linear or spherical linear interpolation (SLERP~\cite{white2016samplinggenerativenetworks}) to merge the weights of different fine-tuned models. Another technique, known as Task Arithmetic~\cite{ilharco2023editing}, enables manipulation of features obtained through fine-tuning by creating \textit{task vectors}, which are derived by subtracting the weights of the original pre-trained model from those of the fine-tuned model. TIES-Merging~\cite{yadav2023tiesmerging} takes this concept further by addressing redundant changes in task vectors and resolving conflicts in parameter signs between multiple task vectors before merging them. This method involves a three-step process: removing small, insignificant parameter changes, resolving sign conflicts between task vectors, merging the adjusted vectors. Additionally, DARE~\cite{yu2024language} proposes randomly dropping some of the changes and rescaling the remaining ones, which can be combined with techniques like Task Arithmetic~\cite{ilharco2023editing} and TIES-Merging~\cite{yadav2023tiesmerging} to enhance the merging process.

Recent research~\cite{akiba2024evolutionaryoptimizationmodelmerging} has also introduced the use of evolutionary algorithms to optimize parameters within TIES-Merging~\cite{yadav2023tiesmerging} with DARE~\cite{yu2024language}. This optimization allows for more granular merging, such as at the level of input/output embedding layers or individual transformer blocks. While previous studies have primarily focused on merging within language models, this research extends the applicability of model merging to vision-language models, demonstrating its effectiveness in this multimodal context.

Most prior studies, except for recent work~\cite{akiba2024evolutionaryoptimizationmodelmerging}, have focused primarily on language models, without extending their methods to multimodal applications. Although recent work~\cite{akiba2024evolutionaryoptimizationmodelmerging} introduced evolutionary algorithm-based optimization and extended model merging to vision-language models, it was limited to natural images and did not extend to domain-specific images beyond natural imagery, such as medical images. Our work is the first to extend evolutionary model merging to the medical domain, specifically for chest X-ray images, demonstrating its effectiveness in this highly specialized context.

%% file: sec/03.tex
\section{JRadiEvo}

JRadiEvo efficiently adapted a vision-language model (VLM) to the non-English medical domain through the evolutionary optimization of TIES-Merging~\cite{yadav2023tiesmerging} combined with DARE~\cite{yu2024language}.

\subsection{Problem setting}

\subsubsection{Vision-language models}

VLM is designed to generate a text response $y$ given an image $x_I$ and accompanying text $x_T$. A typical VLM utilizing a large language model (LLM) is composed of three main components: a vision encoder $\mathcal{M}_V$ that extracts features from the image, a projector $\mathcal{M}_P$ that transforms these image features into the latent space of the LLM, and an LLM component $\mathcal{M}_L$ that generates the output text. The formulation of this process can be expressed as:
\begin{equation}
    y = \mathcal{M}_L(\mathcal{M}_P(\mathcal{M}_V(x_I)), x_T).\label{eq:vlm}
\end{equation}
In this setup, the LLM component $\mathcal{M}_L$ is a pre-trained model that has already acquired strong language capabilities, such as Llama 3~\cite{llama3modelcard}. To adapt it for vision-related tasks, the projector $\mathcal{M}_P$, and optionally the LLM component $\mathcal{M}_L$ are trained or fine-tuned.

\subsubsection{Model merging}
Let $\theta_{\mathrm{init}} \in \mathbb{R}^d$
represent the trainable parameters of the pre-trained LLM, where $d$ is the parameter dimension. Given a set of $K$ tasks 
$\{t_1, t_2, \cdots, t_K\}$, 
the LLM is fine-tuned on these tasks, resulting in a set of fine-tuned parameter vectors 
$\left\{\theta_{\mathrm{ft}}^{t_1}, \theta_{\mathrm{ft}}^{t_2},\cdots, \theta_{\mathrm{ft}}^{t_K}\right\}$. 
Here, each task $t_i$ corresponds to a specific domain or task for fine-tuning the pre-trained LLM, such as vision tasks, medical applications, or adaptation to the Japanese language.

As demonstrated in previous research~\cite{ilharco2023editing}, the task vector for a task $t$ is defined as the difference between the fine-tuned weights for task $t$ and the original pre-trained weights, i.e., the task vector $\tau_t \in \mathbb{R}^d$ is given by:
\[
    \tau_t = \theta_{\mathrm{ft}}^t - \theta_{\mathrm{init}}.
\]
This task vector corresponds to the capabilities acquired through fine-tuning on task $t$. By manipulating these task vectors and merging it with the original weights $\theta_{\mathrm{init}}$, we can merge the capabilities of multiple fine-tuned models without additional training. The details of the merging process we adopted are described in \Sec{merging}.

\subsection{Model merging for JRadiEvo}
\label{sec:merging}

In JRadiEvo, following the approach of previous work~\cite{akiba2024evolutionaryoptimizationmodelmerging}, we optimized TIES-Merging~\cite{yadav2023tiesmerging} combined with DARE~\cite{yu2024language} using an evolutionary algorithm. For the VLM, we also adhered to the strategy used in earlier research~\cite{akiba2024evolutionaryoptimizationmodelmerging} by focusing on the parameters of the LLM component $\mathcal{M}_L$ during the merging process, i.e., $\theta_{\mathrm{ft}}^{t_1}$ represents the parameters of the LLM component $\mathcal{M}_L$ of a VLM fine-tuned on vision-to-text data. Meanwhile,
$\theta_{\mathrm{ft}}^{t_2}, \theta_{\mathrm{ft}}^{t_3},\cdots, \theta_{\mathrm{ft}}^{t_K}$ 
are the parameters of the LLM fine-tuned on text-to-text data, covering tasks $t_2, t_3, \cdots, t_K$ related to medical knowledge or the Japanese language.

The resulting parameters
$\left\{\theta_{\mathrm{ft}}^{t_1}, \theta_{\mathrm{ft}}^{t_2},\cdots, \theta_{\mathrm{ft}}^{t_K}\right\}$ 
were then merged using the corresponding task vectors
$\left\{\tau_{t_1}, \tau_{t_2},\cdots, \tau_{t_K}\right\}$
as follows:

\begin{enumerate}
\item \textbf{DARE}: Following the approach outlined in previous work~\cite{yu2024language}, with $\alpha \in \mathbb{R}$ as the drop rate, the following operation was performed:
    \[
        m^{t} \sim \operatorname{Bernoulli}(\alpha), \quad
        \tilde{\tau}_{\mathrm{DARE}}^{t} = (1 - m^{t})\odot \tau_{\mathrm{ft}}^{t}, \quad
        \tau_{\mathrm{DARE}}^{t} = \tilde{\tau}_{\mathrm{DARE}}^{t} / (1 - \alpha),
    \]
    where $\odot$ denotes element-wise multiplication. This operation randomly drops some of the changes in the task vector $\tau_t$ and rescales the remaining ones.
\item \textbf{TIES-Merging}: Similar to the previous work~\cite{yadav2023tiesmerging}, we removed small, trivial changes in the task vectors, resolved sign conflicts between them, and merged:
    \begin{enumerate}
    \item \textbf{Trim}: To remove insignificant changes, for each task $t$, we created a task vector $\tilde{\tau}_t$ by setting all parameters of a task vector $\tau_{\mathrm{DARE}}^{t}$ to zero except for those absolute values in the top $k_t$ percent.
    \item \textbf{Elect}: To resolve sign conflicts, for each parameter $p \in \{1, 2, \cdots, d\}$, we calculated the sign with greater total movement as $\gamma_m^p = \operatorname{sgn}\left(\sum_{t=1}^{K} \tilde{\tau}_t^p\right)$.
    \item \textbf{Merge}: Finally, for each parameter $p$, we computed the weighted sum of only the parameters from task vectors whose signs matched the aggregated elected sign. Specifically, the merged task vector $\tau_{\mathrm{merged}}$ is given by
    $\tau_{\mathrm{merged}}^p = \frac{1}{|\mathcal{A}_p|}\sum_{t \in \mathcal{A}_p} c_t \tilde{\tau}_t^p$, where $\mathcal{A}_p = \{t \in [n] | \operatorname{sgn}(\tilde{\tau}_t^p) = \gamma_m^p\}$, and $c_t \in \mathbb{R}$ is a weight assigned to each task vector. The merged task vector $\tau_{\mathrm{merged}}$ is then scaled by a scaling parameter $\lambda \in \mathbb{R}$ and added to the initial parameters $\theta_{\mathrm{init}}$ to obtain the final parameters: $\theta_{\mathrm{final}} = \theta_{\mathrm{init}} + \lambda \tau_{\mathrm{merged}}$.
    \end{enumerate}
\item \textbf{Evolutionary optimization}: Following the previous work~\cite{akiba2024evolutionaryoptimizationmodelmerging}, we leveraged an evolutionary algorithm to optimize the parameters of the step above. Specifically, we optimized DARE drop rates $\alpha$, TIES-Merging saved rates $\{k_{t_1}, k_{t_2}, \cdots, k_{t_K}\}$, weight $\{c_{t_1}, c_{t_2}, \cdots, c_{t_K}\}$, and scaling parameters $\lambda$. We treated the entire model as a single layer for the merging process, identical to the approach used in previous research~\cite{akiba2024evolutionaryoptimizationmodelmerging}. We iteratively calculated $\theta_{\mathrm{final}}$ using steps 1 and 2, with $\theta_{\mathrm{final}}$ serving as $\mathcal{M}_L$ in \Eq{vlm}. The parameters were suggested by the evolutionary algorithm to maximize the ROUGE-L~\cite{lin-2004-rouge} score between the generated text $\hat{y}$ from equation \Eq{vlm} and the reference text $y$. This process was repeated, with the evolutionary algorithm continuously refining the parameters to improve the score.
\end{enumerate}

Finally, the text was generated using \Eq{vlm} with the $\theta_{\mathrm{final}}$ obtained and optimized through this process.

%% file: sec/04.tex
\section{Experiments}

\subsection{Experimental setup}
\label{sec:experimental_setup}

\paragraph{Datasets}
\input{tables/dataset}

For our experiments, we used the MIMIC-CXR~\cite{MIMIC-CXR}, a publicly available dataset that consists of 377,110 chest X-ray (CXR) images and 227,835 corresponding English-language radiology reports. The images are provided in both DICOM and JPEG formats~\cite{johnson2019mimiccxrjpglargepubliclyavailable}. 
The dataset is officially split into training, validation, and test sets, containing 368,960, 2,991, and 5,159 images, respectively. The details about the dataset are presented in \Table{dataset}.

Following the previous work~\cite{NEURIPS2018_e0741335,chen-etal-2020-generating}, samples without corresponding reports were excluded from the dataset. Additionally, according to previous research~\cite{NICOLSON2023102633}, for cases where multiple images are associated with a single report, only the first image was used. From the resulting official training set, which included metadata indicating AP (anteroposterior) and PA (posteroanterior) views, we randomly selected 50 samples from both views to create the dataset used in our study. These selected 50 samples were translated into Japanese using GPT-3.5~\cite{NEURIPS2020_1457c0d6}\footnote{We accessed through Azure OpenAI service.\label{fot:azure}}, and the translations were then reviewed and and revised by a human to ensure accuracy.

For the test dataset, we extracted samples from the official test set. As with the training data, only those with corresponding reports were selected, the first image was used for reports associated with multiple images, and we focused exclusively on AP and PA views. These selected samples were translated into Japanese using GPT-3.5~\cite{NEURIPS2020_1457c0d6}\footref{fot:azure}. Note that the training samples were drawn from the official training set and the test samples from the official test set, ensuring no data leakage occurred between the training and test phases.

\paragraph{Evaluation metrics}

Following the previous studies\cite{chen-etal-2020-generating,NICOLSON2023102633,NEURIPS2018_e0741335,Tanida2023InteractiveAE}, we evaluated the generated Japanese radiology reports using BLEU~\cite{papineni2002bleu}, ROUGE-L~\cite{lin-2004-rouge}, and METEOR~\cite{denkowski-lavie-2011-meteor} scores. These metrics are commonly used in machine translation and text generation tasks and provide a comprehensive evaluation of the quality of the generated text.
The generated Japanese reports and the translated reference texts were tokenized using MeCab~\cite{mecab}, a Japanese-specific tokenizer, before calculating the evaluation metrics.

\paragraph{Source models}

To efficiently create a Japanese VLM capable of understanding medical content through evolutionary model merging, we merged a non-medical VLM fine-tuned on vision tasks, two text-to-text LLMs fine-tuned on medical datasets, and another text-to-text LLM fine-tuned on Japanese datasets. Specifically, we used \texttt{Bunny-v1\_1-Llama-3-8B-V}\footnote{\url{https://huggingface.co/BAAI/Bunny-v1_1-Llama-3-8B-V}}~\cite{he2024bunny} as the VLM, \texttt{MMed-Llama-3-8B-EnIns}\footnote{\url{https://huggingface.co/Henrychur/MMed-Llama-3-8B-EnIns}}~\cite{qiu2024building} and \texttt{OpenBioLLM-Llama3-8B}\footnote{\url{https://huggingface.co/aaditya/Llama3-OpenBioLLM-8B}}~\cite{OpenBioLLMs} as the medical models, and \texttt{Llama-3-Swallow-8B-Instruct-v0.1}\footnote{\url{https://huggingface.co/tokyotech-llm/Llama-3-Swallow-8B-Instruct-v0.1}}~\cite{fujii2024continual} as the Japanese language model. All of them are the fine-tunes of the Llama 3~\cite{llama3modelcard}.

\paragraph{Evolutionary optimization}

For the evolutionary algorithm described in \Sec{merging}, we used CMA-ES~\cite{Hansen2006} implemented in Optuna~\cite{akiba2019optuna}, following the previous study~\cite{akiba2024evolutionaryoptimizationmodelmerging}. As hyperparameters for the CMA-ES algorithm, all parameters were initialized to 0.5, with a sigma value of 1/6, and a population size of $4 + \lfloor 3 \ln(n) \rfloor$, where $n$ is the number of parameters. The algorithm was run for 600 iterations, and the best parameters were selected based on the ROUGE-L~\cite{lin-2004-rouge} score. We provided the prompt $x_T$, ``You are a skilled radiologist. Please examine this X-ray image and write a report. Pay attention to any abnormalities in the lungs, heart, or bones. Answer in Japanese,'' written in Japanese.

\subsection{Comparison with leading models from previous study and instruction-tuning approaches}

To evaluate the effectiveness of our evolutionary model merging approach, we compared the results with those obtained by LoRA~\cite{hu2022lora} instruction-tuning the same vision-language model (VLM). Additionally, we compared our results with the performance of leading models from recent research.

\subsubsection{Experimental conditions}

As the base VLM, we used \texttt{Bunny-v1\_1-Llama-3-8B-V}\footnote{\url{https://huggingface.co/BAAI/Bunny-v1_1-Llama-3-8B-V}}~\cite{he2024bunny}, the same model used in the merging process in JRadiEvo, and the fine-tuning.
During this process, the weights of the image encoder $\mathcal{M}_V$ were kept fixed, and we prepared two models: one where only the LLM component $\mathcal{M}_L$ was tuned, and another where both the LLM $\mathcal{M}_L$ and projector $\mathcal{M}_P$ were tuned. Both models were trained using LoRA~\cite{hu2022lora}.
For the LoRA setup, the hyperparameters were set with a decomposition rank $r$ of 8 and a scaling factor $\alpha$ of 16.
The learning rate was was set to $2\times 10^{-4}$ and decayed with a cosine annealing~\cite{loshchilov2017sgdr}.
The model was trained for one epoch with a batch size of 8 and gradient accumulation steps set to 2. 
The training was conducted using a single NVIDIA A100 GPU with 80GB memory.

As for the dataset, we followed the same procedure outlined in \Sec{experimental_setup}, using the training data from MIMIC-CXR~\cite{MIMIC-CXR}. Specifically, samples without corresponding reports were excluded, only the first image was used for reports with multiple images, and we focused on AP and PA view images. From this dataset, 2,000 samples were randomly selected and translated into Japanese using GPT-3.5~\cite{NEURIPS2020_1457c0d6}\footref{fot:azure}. This translated dataset was then used for instruction-tuning.

For comparison with previous studies, we used CheXagent~\cite{chen2024chexagent}, a recent leading model specifically designed for chest X-ray images. CheXagent~\cite{chen2024chexagent}, proposed in 2024, is one of the most recent advancements in the field. It was trained on a large instruction-tuning dataset, which was created by aggregating publicly available data and adding new labels to existing datasets. Since CheXagent is trained in English, reports were first output in English using CheXagent and then translated into Japanese using GPT-4o~\cite{gpt4o}\footref{fot:azure}. This translated Japanese reports were used for comparison.

As another comparison, we used GPT-4o~\cite{gpt4o}\footref{fot:azure}, a high-performance VLM that can output directly in Japanese.

\subsubsection{Results and discussion}

\input{tables/comparison}

The results are shown in \Table{comparison}. Also, an example comparison of generated text from JRadiEvo and the ground truth on the test data is shown in \Table{example}.

\paragraph{Effectiveness of JRadiEvo in generating radiology reports}
We can see from this table that our model, JRadiEvo, achieved the highest scores in both ROUGE-L and METEOR metrics. Given that ROUGE-L is considered the most aligned with human judgment in evaluating generated text, as shown in previous research~\cite{lin-2004-rouge}, this underscores the effectiveness of our application of evolutionary model merging to medical text generation.

\paragraph{Efficient use of limited datasets}
When compared to the latest model, CheXagent~\cite{chen2024chexagent}, JRadiEvo outperformed it across all evaluation metrics. Despite CheXagent being trained on a vast instruction-tuning dataset, JRadiEvo, with only 50 training samples, significantly surpassed it in generating X-ray reports. This demonstrates JRadiEvo's ability to effectively utilize a extremely limited dataset to create a powerful medical foundation model.

\paragraph{Practicality in non-English-speaking regions}
Additionally, while CheXagent~\cite{chen2024chexagent} produces reports in English, our model eliminates the need for an additional translation step by directly generating reports in Japanese. This is particularly important in non-English-speaking medical environments, where neither doctors nor patients may be fluent in English. Requiring translation every time is time-consuming and costly. Thus, JRadiEvo demonstrates potential for practical use in non-English-speaking regions.

\paragraph{Parameter efficiency and local deployment}
In comparison with GPT-4o~\cite{gpt4o}, while it scored higher on BLEU, JRadiEvo outperformed it on METEOR and ROUGE-L, which is more closely aligned with human evaluation. 
This shows that the performance is comparable to, or even surpasses, that of GPT-4o.
Considering that JRadiEvo is a lightweight model with only 8 billion parameters, whereas GPT-4o is significantly larger, this highlights JRadiEvo's impressive parameter efficiency. Furthermore, unlike GPT-4o, which requires API access, JRadiEvo's modest size allows it to be deployed locally in hospitals. This makes JRadiEvo a more practical option for use in privacy- and security-sensitive environments in hospitals.

\paragraph{Superiority over instruction-tuning}
Comparing JRadiEvo with the two models that instruction-tuned the same VLM shows that JRadiEvo outperforms them across all metrics. This highlights the effectiveness of adopting evolutionary model merging over traditional instruction-tuning approaches. Moreover, increasing the amount of data used for instruction-tuning in an attempt to further enhance the model can lead to catastrophic forgetting~\cite{MCCLOSKEY1989109,kotha2024understanding,luo2024empiricalstudycatastrophicforgetting}. In our experiments, instruction-tuning with 2,000 data points yielded good results, but when we increased the dataset to 10,000 data points, catastrophic forgetting occurred, severely impairing the language functionality and rendering the model unusable. As previous research~\cite{wang2023lora} has indicated, instruction tuning with LoRA~\cite{hu2022lora} in Japanese is not effective for relatively small models, a finding that our results also confirm. This suggests that similar challenges may arise in other non-English languages as well. In contrast, JRadiEvo leveraged evolutionary model merging to successfully adapt the source VLM to the medical domain without experiencing catastrophic forgetting. This highlights the potential of model merging as a viable approach for domain adaptation in non-English languages, especially for smaller models.

\input{tables/example}

\subsection{Analysis of merged LLM contributions}

\begin{figure}[tb]
    \centering
    \includegraphics[width=115mm]{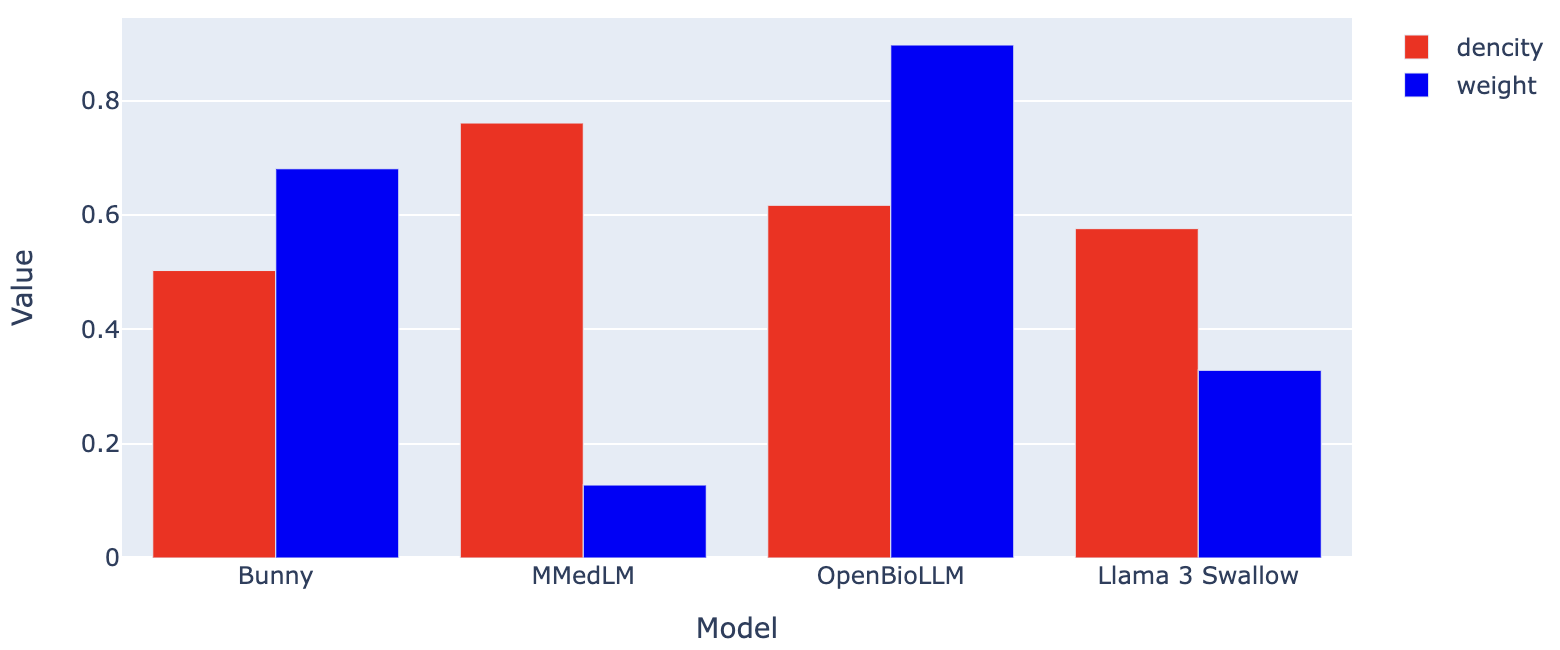}
    \caption{Evolved configurations for each models.}
    \label{fig:merged_models}
\end{figure}

To investigate the contributions of the merged LLMs, we compared the density and weight parameters after optimization, following the approach outlined in previous research~\cite{akiba2024evolutionaryoptimizationmodelmerging}. Specifically, we analyzed the retained percentage $\{k_{t_1}, k_{t_2}, \cdots, k_{t_K}\}$ and the merging weight $\{c_{t_1}, c_{t_2}, \cdots, c_{t_K}\}$ described in \Sec{merging}. The results of this comparison are presented in \Fig{merged_models}.

As shown in \Fig{merged_models}, the weight for OpenBioLLM~\cite{OpenBioLLMs} is significantly higher, and its density is also relatively high. This suggests that the medical knowledge embedded in OpenBioLLM was crucial for adapting the non-medical VLM to the medical domain. In contrast, while MMedLM~\cite{qiu2024building} has a high density, its weight is much lower, indicating that its contribution was less significant. This suggests that, in this setup, OpenBioLLM's medical knowledge was primarily utilized, potentially rendering MMedLM less influential in the model merging process.

Additionally, when examining the density and weight of Llama 3 Swallow~\cite{fujii2024continual}, the LLM enhanced for Japanese language proficiency, we see that it was utilized to some extent, though not as heavily as OpenBioLLM. This suggests that the original VLM and LLMs had a some capacity for handling Japanese, albeit imperfectly, and the lack of medical knowledge was a more significant limitation. However, unlike MMedLM, which saw a dramatic reduction in weight, Llama 3 Swallow's contribution was not negligible, indicating that its role was still necessary. In fact, when asked directly in Japanese, the other VLM and LLMs could generate responses that, while somewhat awkward and unnatural from a native speaker's perspective, were still intelligible.

%% file: tables/dataset.tex
\begin{table}[t]
    \centering
    \caption{
        The statistics of the MIMIC-CXR dataset~\cite{MIMIC-CXR}, showing the number of images, reports, and patients in each split.
    }
    \label{tbl:dataset}
    \scalebox{0.95}{
        \begin{tabular}{lrrr}
            \toprule
            & Train  & Valid & Test \\
            \midrule
            Image & 368,960 & 2,991 & 5,159 \\
            Report & 222,758 & 1,808 & 3,269 \\
            Patient & 64,586 & 500 & 293 \\
            \bottomrule
        \end{tabular}
    }
\end{table}

%% file: tables/comparison.tex
\begin{table}[t]
  \centering
  \caption{
    Comparison of JRadiEvo with leading models from recent research and instruction-tuned approaches. \textbf{Bold} and \underline{underlined} scores are the best and worst in each metric, respectively.
  }
  \label{tbl:comparison}
  \scalebox{0.87}{
      \begin{tabular}{lcccccc}
          \toprule
                                  & BLEU-1 & BLEU-2 & BLEU-3 & BLEU-4 & ROUGE-L & METEOR \\
          \midrule
          JRadiEvo (ours)                            & 0.376        & 0.301  & 0.241  & 0.179  &\textbf{0.212}&\textbf{0.191}\\
          LLM instruction-tuned        & 0.342     & 0.241  & 0.185  & 0.133  &\underline{0.143}&\underline{0.142}\\
          LLM + projector instruction-tuned & 0.345  & 0.245 & 0.189 & 0.136 & 0.147 & 0.146\\
          \midrule
          CheXagent~\cite{chen2024chexagent}         &\underline{0.221}&\underline{0.181}&\underline{0.149}&\underline{0.119}& 0.199  & 0.159  \\
          GPT-4o~\cite{gpt4o}                        & \textbf{0.432}     &\textbf{0.333}&\textbf{0.257}&\textbf{0.184} & 0.176  & 0.188  \\
          \bottomrule
      \end{tabular}
  }
\end{table}

%% file: tables/example.tex
\newcommand{\pwidth}{85mm}
\newcommand{\textsize}{\fontsize{6.8pt}{6.7pt}\selectfont}

\begin{table}[t]
    \centering
    \caption{Example comparison of generated text from JRadiEvo and the ground truth on the test data. (For reference, the English translation by GPT-4o is shown in parentheses.)}
    \label{tbl:example}
    \begin{tabular}{c}
        \toprule
        \textit{\small Example 1}\\
        \midrule
        \begin{minipage}{30mm}
            \includegraphics[width=25mm]{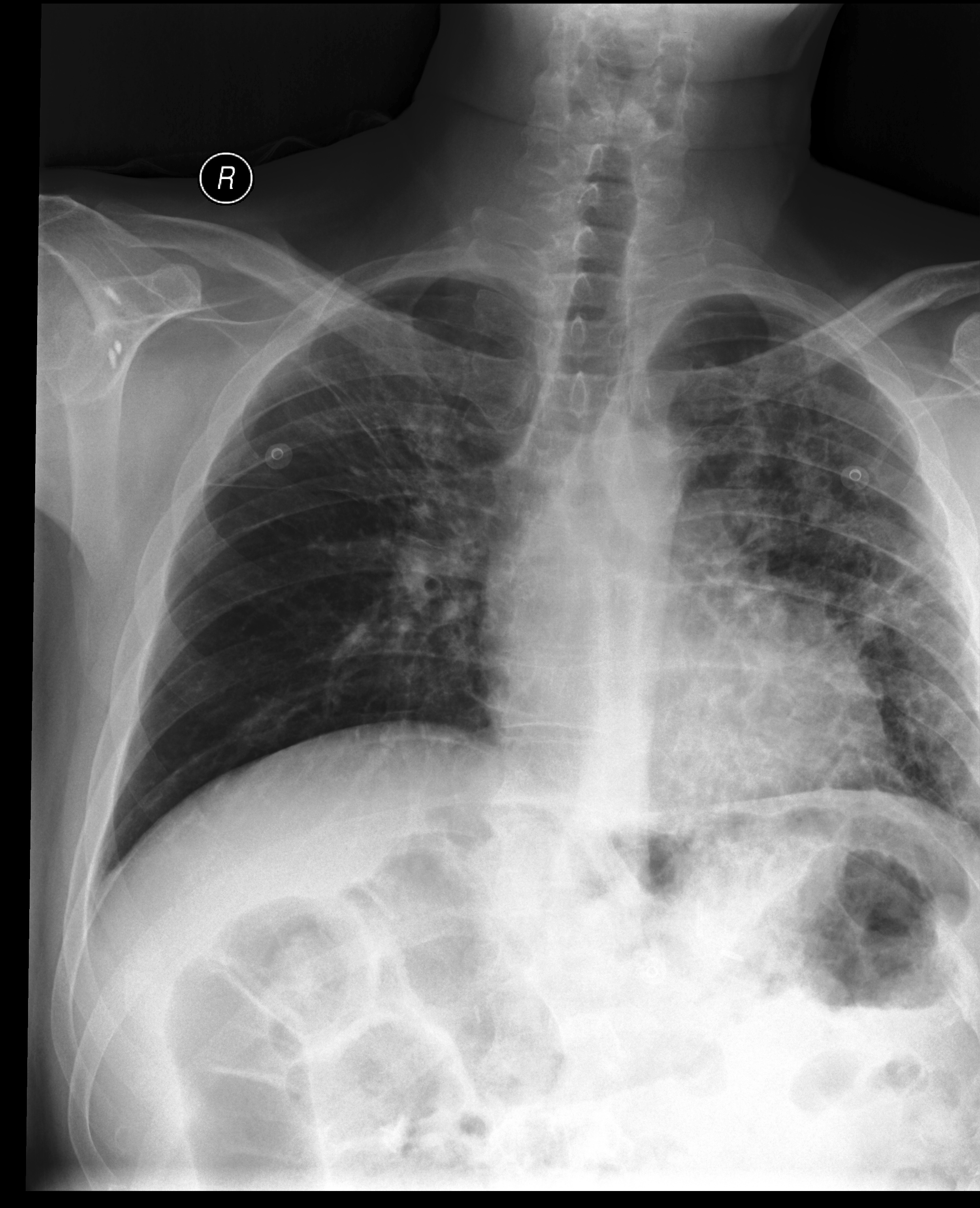}
        \end{minipage}
        \hspace{-5mm}
        \begin{minipage}{108mm}
            \begin{tabular}{m{12mm}l}
                {\footnotesize Generated}
                &\Ja{\textsize
                    \begin{tabular}{p{\pwidth}}
                    レントゲン写真を読影した結果、肺野に多数の小さな
                    白斑が見られます。これは肺炎の可能性があります。
                    心臓の形状は正常です。骨には異常は見られません。\\
                    (The X-ray examination shows multiple small white spots in the lung fields, which may indicate pneumonia. The shape of the heart is normal, and no abnormalities are observed in the bones.)
                    \end{tabular}
                }\\
                \midrule
                {\footnotesize Ground~truth}
                &\Ja{\textsize
                \begin{tabular}{p{\pwidth}}
                    肺容量は低下している。左肺上葉と比較し、右上葉により強く瘢痕性、慢性の線維化が見られる。左肺中部に線状影が見られ、左下葉にも軽度見られる。胸水はない。心臓、縦郭の輪郭は正常である。\\
                    (The lung volume is reduced. Compared to the left upper lobe, the right upper lobe shows more pronounced scarring and chronic fibrosis. Linear opacities are observed in the mid-left lung, and mild findings are also present in the left lower lobe. There is no pleural effusion. The contours of the heart and mediastinum are normal.)
                    \end{tabular}
                }\\
            \end{tabular}\\
        \end{minipage}\\
        \toprule
        \textit{\small Example 2}\\
        \midrule
        \begin{minipage}{30mm}
            \includegraphics[width=25mm]{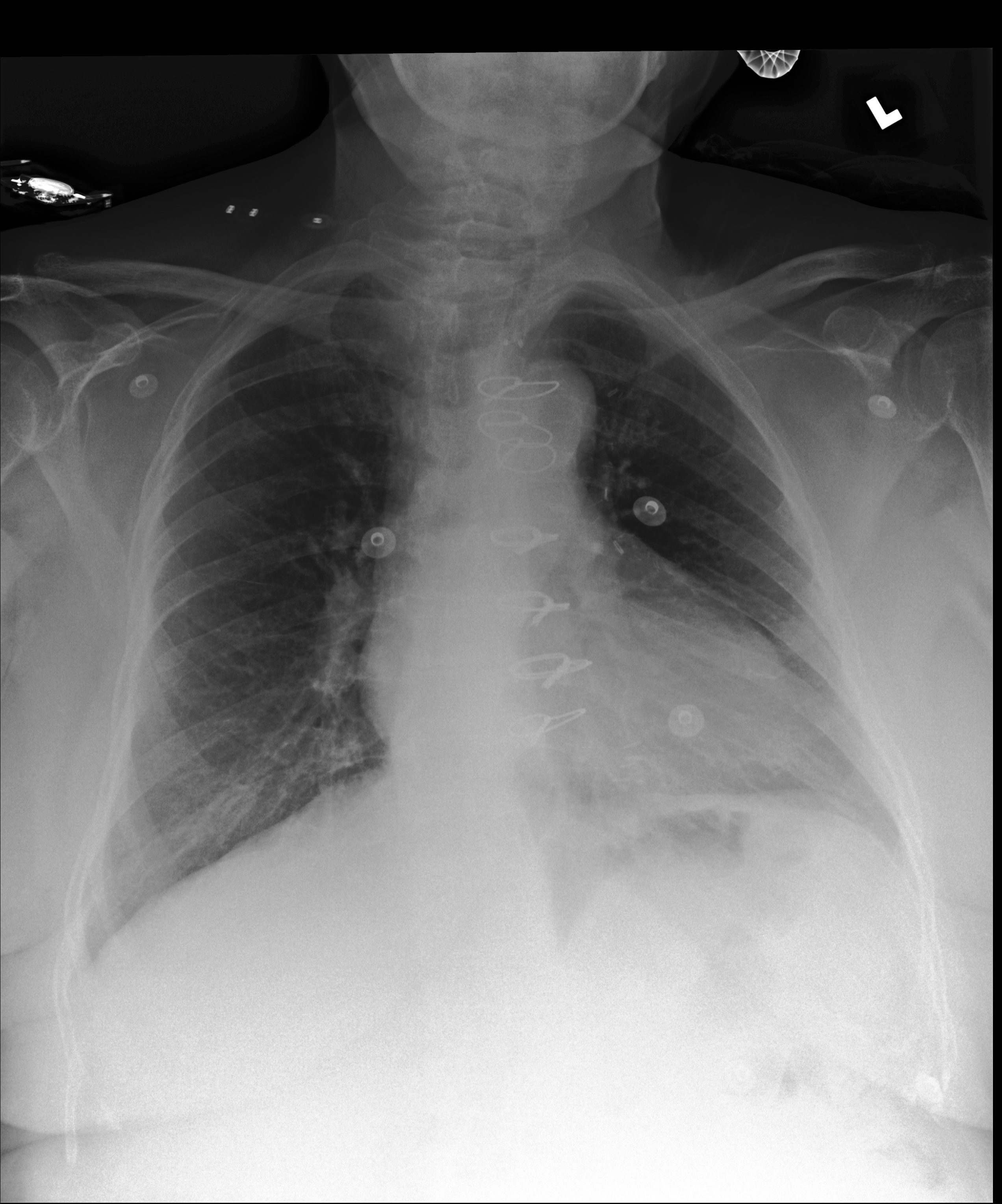}
        \end{minipage}
        \hspace{-5mm}
        \begin{minipage}{108mm}
            \begin{tabular}{m{12mm}l}
                {\footnotesize Generated}
                &\Ja{\textsize
                    \begin{tabular}{p{\pwidth}}
                        レントゲン写真を読影した結果、心臓の左室が大きく、右室が小さいことがわかりました。肺野は均一に空気を含み、骨は正常です。心臓の左室の異常は心不全の可能性を示唆しています。\\
                    (The X-ray examination revealed that the left ventricle of the heart is enlarged, while the right ventricle is small. The lung fields are uniformly aerated, and the bones are normal. The abnormality in the left ventricle suggests the possibility of heart failure.)
                    \end{tabular}
                }\\
                \midrule
                {\footnotesize Ground~truth}
                &\Ja{\textsize
                \begin{tabular}{p{\pwidth}}
                    胸骨正中切開のワイヤーと縦郭のクリップが見られる。肺野は清明で、浸潤影や胸水、気胸はない。心臓や縦郭陰影は正常。骨病変はない。右横隔膜下にフリーエアはない。\\
                    (Wires from a median sternotomy and clips in the mediastinum are visible. The lung fields are clear, with no infiltrates, pleural effusion, or pneumothorax. The cardiac and mediastinal silhouettes are normal. There are no bone lesions. No free air is seen under the right diaphragm.)
                    \end{tabular}
                }\\
            \end{tabular}\\
        \end{minipage}\\
        \toprule
        \textit{\small Example 3}\\
        \midrule
        \begin{minipage}{30mm}
            \includegraphics[width=25mm]{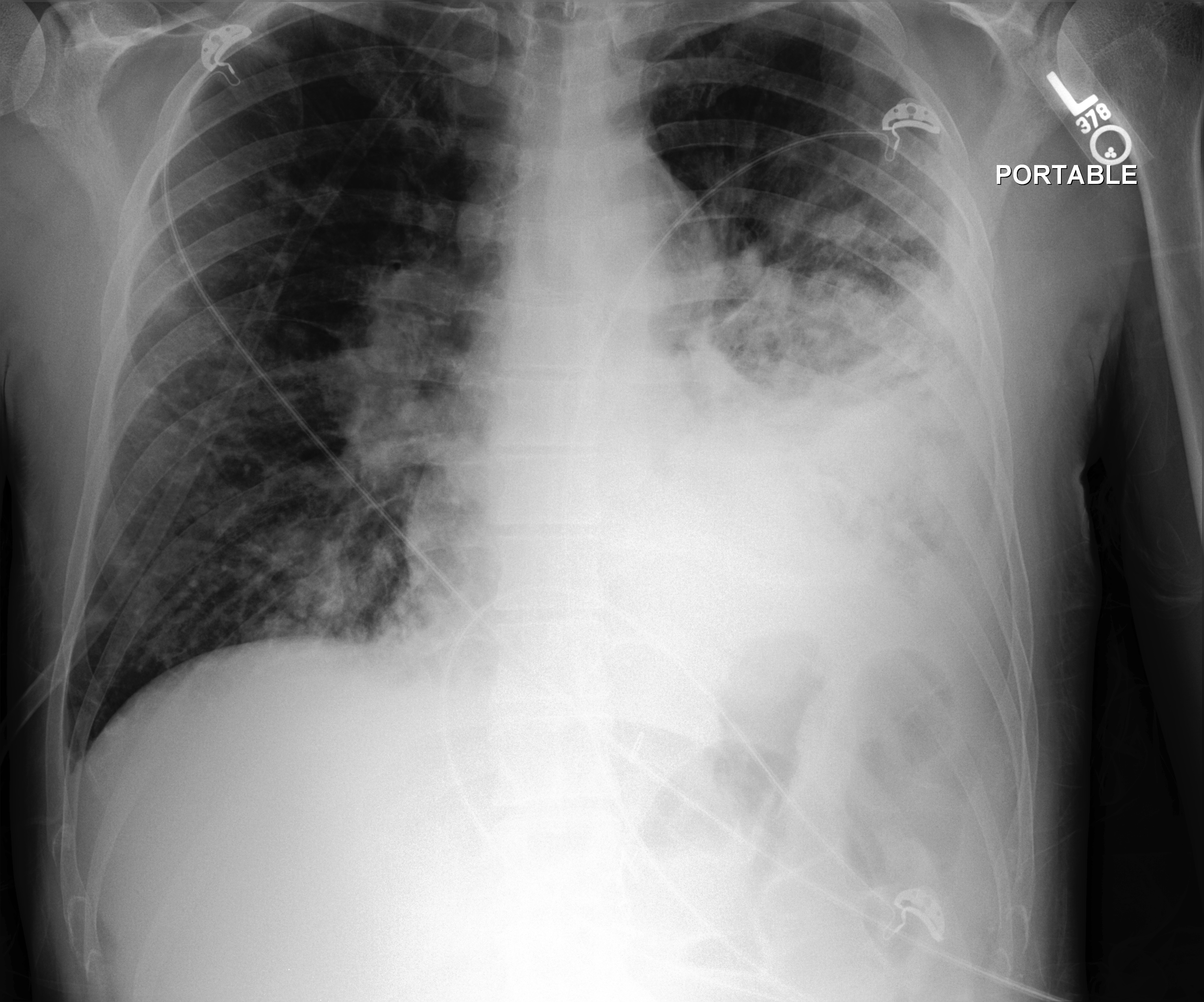}
        \end{minipage}
        \hspace{-5mm}
        \begin{minipage}{108mm}
            \begin{tabular}{m{12mm}l}
                {\footnotesize Generated}
                &\Ja{\textsize
                    \begin{tabular}{p{\pwidth}}
                        レントゲン写真を読影した結果、左肺下葉に不規則な影が見られます。心臓の形状は正常です。骨には異常は見られません。左肺下葉の影は肺炎や肺癌などの可能性があります。詳細な検査が必要です。\\
                    (The X-ray examination shows an irregular opacity in the left lower lobe. The shape of the heart is normal, and no abnormalities are observed in the bones. The opacity in the left lower lobe may suggest possibilities such as pneumonia or lung cancer. Further detailed examination is required.)
                    \end{tabular}
                }\\
                \midrule
                {\footnotesize Ground~truth}
                &\Ja{\textsize
                \begin{tabular}{p{\pwidth}}
                    ポータブル撮影画像。左肺下部の病変が拡大しており、胸水と浸潤影を疑う。肺野に散在する結節影は悪性腫瘍の転移を疑う。心陰影は不明瞭で心拡大の評価は困難。骨病変は指摘できず。\\
                    (Portable imaging shows an enlarged lesion in the lower left lung, raising suspicion of pleural effusion and infiltrates. Scattered nodules in the lung fields suggest possible metastasis of a malignant tumor. The cardiac silhouette is unclear, making it difficult to assess for cardiomegaly. No bone lesions are noted.)
                    \end{tabular}
                }\\
            \end{tabular}\\
        \end{minipage}\\
        \bottomrule
    \end{tabular}
\end{table}